%% file: l4dc2021-sample.tex
\documentclass[12pt,final]{l4dc2021}

\title[Uncertainty-aware Safe Exploratory Planning using GP and NCCM]{Uncertainty-aware Safe Exploratory Planning using Gaussian Process and Neural Control Contraction Metric}
\usepackage{times}

\makeatletter
\def\set@curr@file#1{\def\@curr@file{#1}} %
\makeatother
\usepackage[load-configurations=version-1]{siunitx} %
\input{NewMacros}
\usepackage[T1]{fontenc}
\usepackage{amssymb,amsmath}
\usepackage[latin5]{inputenc}
\usepackage{graphicx,times,latexsym}
\usepackage{psfrag}
\usepackage{setspace}
\usepackage{color}
\usepackage{url}
\usepackage{wrapfig}
\usepackage{caption}
\usepackage{comment}
\usepackage{paralist}

\usepackage{mathtools}
\mathtoolsset{showonlyrefs=true}

\newcommand{\oprocendsymbol}{\hbox{$\bullet$}}
\newcommand{\oprocend}{\relax\ifmmode\else\unskip\hfill\fi\oprocendsymbol}

\newcommand{\te}{\mathcal{E}}

\newcommand{\extended}{}

\author{%
 \Name{Dawei Sun} \Email{daweis2@illinois.edu}\\
 \addr University of Illinois Urbana-Champaign, Urbana, IL 61801, USA
 \AND
 \Name{Mohammad Javad Khojasteh} \Email{mkhojast@mit.edu}\\
 \addr Massachusetts Institute of Technology, Cambridge, MA 02139, USA%
 \AND
 \Name{Shubhanshu Shekhar} \Email{shshekha@eng.ucsd.edu}\\
 \addr University of California San Diego, La Jolla, CA 92093, USA%
 \AND
 \Name{Chuchu Fan} \Email{chuchu@mit.edu}\\
 \addr Massachusetts Institute of Technology, Cambridge, MA 02139, USA%
}

\input{sym.tex}

\input{macros}

\begin{document}

\maketitle

\begin{abstract}
In this paper, we consider the problem of using a robot to explore an environment with an unknown, state-dependent disturbance function while avoiding some forbidden areas. The goal of the robot is to safely collect observations of the disturbance and construct an accurate estimate of the underlying disturbance function. We use Gaussian Process~(GP) to get an estimate of the disturbance from data with a high-confidence bound on the regression error. Furthermore, we use neural Contraction Metrics to derive a tracking controller and the corresponding high-confidence uncertainty tube around the nominal trajectory planned for the robot, based on the estimate of the disturbance. From the robustness of the Contraction Metric, error bound can be pre-computed and used by the motion planner such that the actual trajectory is guaranteed to be safe. As the robot collects more and more observations along its trajectory, the estimate of the disturbance becomes more and more accurate, which in turn improves the performance of the tracking controller and enlarges the free space that the robot can safely explore. We evaluate the proposed method using a carefully designed environment with a ground vehicle. Results show that with the proposed method the robot can thoroughly explore the environment safely and quickly.
\end{abstract}


\section{Introduction}

In the past few years, there has been an increasing interest in combining learning-based system identification and control theoretic techniques to accomplish complex tasks and control objectives~\citep{deisenroth2011pilco,Dean2019,sarkar2019finite,coulson2019data,chen2018approximating,liu2020robust,fan2020deep,chowdhary2014bayesian,jagtap2020control,levine2016end,pan2018agile,kahn2020badgr,thananjeyan2020safety,srinivasan2020synthesis,wabersich2020bayesian,wabersich2020performance}. Such a combination has shown to be able to reconcile the advantages of (deep) learned models which better represent data but are hard to be analyzed, and control techniques that are proven to work robustly but only on well-modeled control systems. Following this line of work, we study the problem of  motion planning for robots to better learn the model uncertainties, while maintaining safety during the exploration process.

\begin{wrapfigure}{r}{0.35\textwidth}
    \captionsetup{format=plain}
    \centering
    \includegraphics[width=0.35\textwidth]{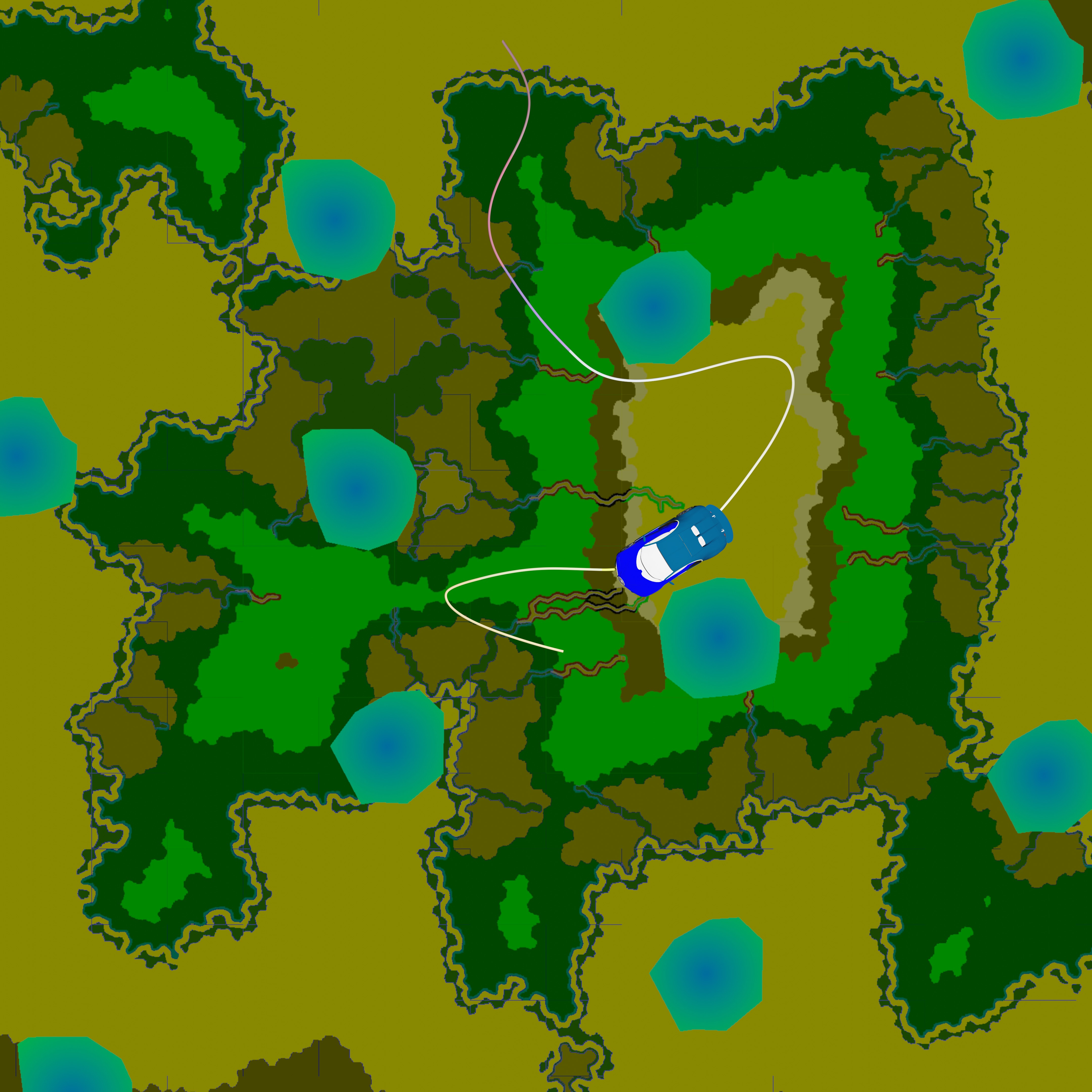}
    \caption{\footnotesize Motivating scenario: A ground vehicle needs to safely explore unseen environments to learn the effect of different terrains (in different colors) on its dynamics. The light blue regions are pools that the vehicle should avoid.}
    \label{fig:scene}
\end{wrapfigure}

Consider the motivating example in Figure~\ref{fig:scene}. The dynamics of a ground vehicle contain a disturbance term, which is an unknown function of the current position. For example, the friction factor will be different while the vehicle is driving on sand or grass. There are pools that the vehicle should avoid. To learn an accurate model of the vehicle, we have to safely drive the vehicle to every part of the environment and collect data about the friction while remaining safe. Similarly, safe exploratory planning is also a key yet challenging problem in many engineering domains such as Mars rover exploration as in~\citep{ono2018mars,ahmadi2020risk,strader2020perception} and delivery drones as in~\citep{cao2017gaussian,berkenkamp2015safe}.

We propose a novel framework to solve such uncertainty-aware safe exploratory problems by combining neural Control Contraction Metric (CCM)~\citep[cf.][]{sun2020learning} 
with Gaussian Process (GP).
Let us use the above scenario as an example. The vehicle has dynamics $\dot{\bfx} = f(\bfx) + B(\bfx)\bfu + d(\bfx)$ where $d$ is the unknown but bounded model error. The robot aims to approximate $d$ with a GP model $\hat{d}$. Initially, the vehicle is only aware of its immediate surrounding environment. That is, it gets an initial guess of $\hat{d}$ and knows an upper bound on the variance of the estimation error $\|\hat{d} - d\|$ in a ball of radius $\rho$ around itself. The algorithm then learns a robust control law using CCM such that in the $\rho$-ball, the vehicle can track any desired path $\bfx^*(t)$ with a tracking error $\te$. Therefore, a reference path $\bfx^*(t)$ is safe, if it is guaranteed to be at least $\te$ away from the forbidden areas. Then, at each time step, the vehicle will plan a safe path within the $\rho$-ball with the goal of obtaining more information about $d$. The algorithm will collect samples along the traveled path to continue updating the model $\hat{d}$, which in turn improves the performance of the tracking controller and characterizes more free space as safe to explore.
This process will terminate when the free space has been fully explored in the sense that the estimation error $\|\hat{d} - d\|$ is uniformly below a threshold.

We evaluate the proposed method in the scenario as shown in Figure~\ref{fig:scene}. We compare the proposed method with a baseline method where the error $d$ is not learned on the fly but pre-estimated with hand-crafted bounds. Results show that with the proposed method the agent spends a shorter time 
exploring the environment and results in fewer collisions, which demonstrates the sample efficiency and safety guarantee of the proposed method. Moreover, by combining piece-wise linear paths and learned tracking controllers our method deals with nonlinear dynamics efficiently.

The major contributions of this paper are as follows: Firstly, we propose a framework for combining the GP model with neural contraction metric to safely collect data in an unknown environment. Secondly, we investigate the sample complexity of the GP regression and use the estimation variance of GP to determine the next point to explore, which improves the sample efficiency.
Thirdly, we derive the criteria for determining when to update the tracking controller, which reduces unnecessary computation.

\paragraph{Related work} 
Safe exploration has been studied in an extensive set of publications. Here we only mention a non-exhaustive list of related work.  \cite{liu2020robust} used neural networks to learn the residual dynamics from data and use statistical learning theory to get a bound on the control performance. In \citep{nakka2020chance} the learned dynamics is projected into a finite-dimensional space using generalized polynomial chaos, and the trajectory planning problem is written as a convex optimization problem based on the approximated dynamics. \cite{pravitra2020l1} used model predictive path integral control (MPPI) for motion planning, and used $L1$ adaptive control for handling the potential mismatch between the nominal and true dynamics. \cite{koller2018learning} and \cite{wabersich2020bayesian,wabersich2020performance} propose learning-based model predictive control (MPC) schemes that provide high-probability safety guarantees throughout the learning process using GPs. In MPC, nonlinear dynamics show up as constraints of the optimization problem, which reduce the efficiency of such methods for robots with complicated dynamics. 
\cite{berkenkamp2016safe} used GP to learn the unknown part of the dynamics and used Lyapunov functions to determine a region of attraction (ROA). With these guarantees, they provided an algorithm to actively and safely explore the state space to expand the ROA. \cite{berkenkamp2016safe22}
optimized the parameters of a controller while ensuring safety by modeling the underlying performance measure as a GP.

\input{setup}

\input{sample_complexity}

\input{tracking_error}

\input{algo}

\input{experiments}

\section{Discussion and Future Work}
\label{sec:conc2}

In this paper, we consider the problem of using a robot to safely explore an unknown environment and propose a framework where GP and contraction metric are combined to drive the robot efficiently and safely in the environment. Results on a ground vehicle model verify the efficiency of the proposed safe exploration framework. There are several interesting directions for future research. 

\begin{itemize}\itemsep0em
\vspace{-0.8em}
    \item 
We developed an independent GP regression for each coordinate.
For cases where strong correlations exist between components, we could employ Matrix-Variate GP, as in existing works such as \citep{khojasteh2019probabilistic, louizos2016structured, cheng2020safe}.

\item
In this work, we assume the unknown part of the dynamics is a sample from GP. Alternatively, depending on specific applications and the available prior knowledge, it may be more suitable to apply other estimation techniques such as random forests, neural networks or counter-example guided learning~\citep{chen2020counter}.

\item
In this work, we plan the agent action toward the point with the highest estimate variance~\eqref{eq:next_point} and empirically showed its benefits.
An important question to investigate for future work is whether there exist planning strategies that can provably improve upon our method. 
Ideas from the  literature on active learning~\citep{buisson2020actively,capone2020localized,lew2020safe,nakka2020chance} may be useful in designing such optimal strategies. 
\end{itemize}

\noindent\textbf{Acknowledgments} The authors acknowledge support from the DARPA Assured Autonomy under contract FA8750-19-C-0089 and from the Defense Science and Technology Agency in Singapore. The views, opinions, and/or findings expressed are those of the authors and should not be interpreted as representing the official views or policies of the Department of Defense, the U.S. Government, DSTA Singapore, or the Singapore Government.

\bibliography{mybib}

\ifx\extended\undefined
\else
\newpage 
\appendix 

\section{Proof of Theorem~\ref{prop:sample_complexity}}
\label{appendix:proof_sample_complexity}
\input{appendix_proof_sample_complexity}

\section{Proof of Theorem~\ref{thm:retraining} } 
\label{appendix:proof_retraining}
\input{appendix_proof_retraining}

\section{More Experimental Results}
\label{appendix:moreexp}
\input{appendix_moreexp}
\fi
\end{document}

%% file: NewMacros.tex
\newcommand*{\lp}{\left(}
\newcommand*{\rp}{\right)}

\newcommand{\mc}[1]{\mathcal{#1}}
\newcommand{\mbb}[1]{\mathbb{#1}} 
\newcommand{\tbf}[1]{\textbf{#1}}

\newtheorem{assumption}{Assumption}

\newcommand*{\cl}{\underline{c}}
\newcommand{\ball}{\B}

%% file: sym.tex
\newcommand{\bfg}{\mathbf{g}}

\newcommand{\bfo}{\mathbf{o}}

\newcommand{\bfu}{\mathbf{u}}
\newcommand{\bfv}{\mathbf{v}}

\newcommand{\bfx}{\mathbf{x}}
\newcommand{\bfy}{\mathbf{y}}

%% file: macros.tex
\usepackage{scalerel,stackengine}
\stackMath

\def\B{{\cal B}} %

\newcommand{\nnnum}[1]{\relax\ifmmode 
	{\mathbb #1}_{\geq 0} \else ${\mathbb #1}_{\geq 0}$
	\fi}
\newcommand{\npnum}[1]{\relax\ifmmode 
	{\mathbb #1}_{\leq 0} \else ${\mathbb #1}_{\leq 0}$
	\fi}
\newcommand{\pnum}[1]{\relax\ifmmode 
	{\mathbb #1}_{> 0} \else ${\mathbb #1}_{> 0}$
	\fi}
\newcommand{\nnum}[1]{\relax\ifmmode 
	{\mathbb #1}_{< 0} \else ${\mathbb #1}_{< 0}$
	\fi}
\newcommand{\plnum}[1]{\relax\ifmmode 
	{\mathbb #1}_{+} \else ${\mathbb #1}_{+}$
	\fi}
\newcommand{\nenum}[1]{\relax\ifmmode 
	{\mathbb #1}_{-} \else ${\mathbb #1}_{-}$
	\fi}

\newcommand{\reals}{{\mathbb{R}}}                    %
\newcommand{\nnreals}{{\nnnum R}}                    %

\newcommand\reallywidehat[1]{%
\savestack{\tmpbox}{\stretchto{%
  \scaleto{%
    \scalerel*[\widthof{\ensuremath{#1}}]{\kern-.6pt\bigwedge\kern-.6pt}%
    {\rule[-\textheight/2]{1ex}{\textheight}}%
  }{\textheight}%
}{0.5ex}}%
\stackon[1pt]{#1}{\tmpbox}%
}
\newcommand{\sym}[1]{\mathtt{sym}\left(#1\right)}

\newcommand{\psith}{\psi_{\mathtt{th}}}
\newcommand{\thr}{\mathtt{thr}}

%% file: setup.tex
\section{Problem setup and notations}
\label{sec:prelim}
We denote by $\reals$ and $\nnreals$ the set of real and non-negative real numbers respectively. For a symmetric matrix $A \in \reals^{n\times n}$, the notation $A \succ 0$ means $A$ is positive definite.
For a matrix-valued function $M(\bfx):\reals^n \mapsto \reals^{n \times n}$, its element-wise Lie derivative along 
a vector $\bfv \in \reals^{n}$ is $\partial_\bfv M := \sum_{i} \bfv^{(i)} \frac{\partial M}{\partial \bfx^{(i)}}$. Unless otherwise stated, $\bfx^{(i)}$ denotes the $i$-th element of vector $\bfx$. For $A \in \reals^{n \times n}$, we denote $A + A^\intercal$ by $\sym{A}$. The ball centered at $\bfx$ with radius $\rho$ is denoted by $\ball(\bfx, \rho)$.

We consider the problem of exploring an environment with unknown state-dependent disturbances and known obstacles. Assume that $\mathcal{X} \subseteq \reals^n$ is the state space, $\mathcal{U} \subseteq \reals^m$ is the input space, $\mathcal{D} \subseteq \reals^n$ is the domain of disturbances, and $\mathcal{O} \subset \mathcal{X}$ is the region containing the obstacles. Let $\bfx(t) \in \mathcal{X}$ be the state of the agent; then the dynamics is given by
\begin{align}
\label{dynm1-7}
    \dot{\bfx}=f\left(\bfx(t)\right) + B\left(\bfx(t)\right)\bfu(t) + d(\bfx(t)),
\end{align}
where dynamics functions $f: \mathcal{X} \mapsto \reals^n$, $B: \mathcal{X} \mapsto \reals^{n \times m}$ are smooth, $\bfu: \nnreals \mapsto \mathcal{U}$ is the control input, and $d: \mathcal{X} \mapsto \mathcal{D}$ is a disturbance function. The functions $f$ and $B$ are assumed to be \textit{known}, whereas $d$ represents the \textit{unknown} part of the dynamics, caused by discrepancies between the model and the real dynamics or by disturbances in the environment, such as drag or friction.

We assume that the agent can observe the disturbance $d(\bfx)$ after it has visited a small neighborhood around state $\bfx$, that is, it only collects disturbance observations around its trajectory. The observations are noisy with i.i.d. additive Gaussian noise with zero mean and covariance $s^2I_{n}$. 
The goal of the agent is to safely explore the environment to establish an accurate estimate $\hat{d}(\cdot)$ of the disturbance map $d(\cdot)$. At the same time, it should make use of the current estimate to explore the environment while avoiding the obstacles. 
Let $e(\bfx) = d(\bfx) - \hat{d}(\bfx)$ be the estimation error.
Formally, the overall goal is to find an estimate $\hat{d}$ such that $\|e(\bfx)\|_2 \leq \psith$ for all $\bfx \in \mathcal{X}$ in the free-space, and some given threshold $\psith > 0$, while ensuring safety during exploration.

To derive analytical results, we need to limit the class of possible uncertainty map $d$. In particular, we work in a Bayesian framework and assume that $d$  is a sample from a multivariate Gaussian process  with zero mean and
known  covariance function (or kernel) $\mathcal{K}(\cdot, \cdot)$~\citep[cf.][]{srinivas2012information,lederer2019uniform}.  
The choice of the kernel is problem dependent; see, e.g., \citep{williams2006gaussian} for a review of common kernel choices. In addition we assume that the kernel $\mathcal{K}$ satisfies the following properties:
\begin{assumption}
\label{assump:covariance}
\begin{inparaenum}[(i)]
\item $\mathcal{K}$ is \emph{isotropic}, i.e., $\mc{K}(\bfx, \bfy)$ depends on $\bfx$ and $\bfy$ only through $\|\bfx-\bfy\|$ and hence in the sequel we will also overload the notation and use $\mc{K}(\|\bfx-\bfy\|)$ to denote $\mc{K}(\bfx,\bfy)$;
\item There exist constants $C_K>0$ and $\omega \in (0,1]$~(depending on $\mc{K}$) such that
we have $\sqrt{2(\mc{K}(0) - \mc{K}(r))} \leq C_K r^\omega$ for all $r>0$. This condition is satisfied for most of the commonly used covariance functions such as Squared-Exponential~(SE) and  Mat\'ern kernels (with half-integer smoothness) as noted by \cite{shekhar2018gaussian};
\item There exist constants $a_1, a_2, L >0$, such that \\$\mbb{P} \lp \{ \sup_{\bfx \in \mc{X}}\,|\partial d^{(j)}(\bfx)/\partial \bfx^{(j)}| < L \}\rp \geq 1 - a_1 n e^{-L^2/a_2^2}$ for $j = 1,\ldots,n$. Note that this assumption was employed in \citep{srinivas2009gaussian} to apply the GP based analysis to continuous domains $\mc{X}$.
\end{inparaenum}
 \end{assumption}

\paragraph{Overview of the method.} The proposed method consists of three major components.
\begin{inparaenum}[(i)]
\item Gaussian Processes (GP) are used to learn the disturbance from observations and give the corresponding high-probability bound on the estimation error, which will be elaborated in Sec.~\ref{sec:GP};
\item With the estimate of the disturbance, we apply the method proposed by~\cite{sun2020learning} to learn a tracking controller for the approximated dynamics. Using this controller, the system can track any nominal trajectory with bounded tracking error, which will be elaborated in Sec.~\ref{sec:C3M};
\item An uncertainty-aware data acquisition algorithm is used to ensure that the agent always visits the most informative points such that the estimation error can be efficiently reduced as the agent collects data around its trajectory. Also, a simple planning strategy is used to plan nominal trajectories in the environment considering the pre-computed tracking error bound such that the motion of the agent is guaranteed to be safe. The overall exploration algorithm will be shown in Sec.~\ref{sec:algo}.
\end{inparaenum}

%% file: sample_complexity.tex
\section{Gaussian process regression and sample complexity}
\label{sec:GP}
We use GP as our Bayesian inference tool to estimate state-dependent disturbances $d$. Following~\citep{berkenkamp2017safe} we develop a unidimensional GP regression for each dimension $d^{(i)}$, where $i=1,\ldots,n$. 
 Recall that the observations are disturbed with i.i.d. additive Gaussian noise with  zero mean and covariance $s^2I_{n}$. The training observations, for the $i$-th  coordinate, at the sampling points $\bfx_{[N]}:=[\bfx_1, \ldots,\bfx_N]^\intercal$, are denoted by $\textbf{y}_{i,[N]},$ which is the noisy version of the vector $[d^{(i)}(\bfx_1), \ldots, d^{(i)}(\bfx_N)]^{\intercal}$. Let $\kappa_i$ be the kernel function for the $i$-th coordinate. The posterior distribution is again Gaussian and can be computed at the query test point $\bfx_*$,  as follows ~\citep[cf.][]{williams2006gaussian}.  
\begin{align}
\label{eq:GP_inf}
    & d^{(i)}(\bfx_*) \sim \mathcal{N} \big(\mu_N^{(i)}(\bfx_*),\; \sigma_N^{(i)}(\bfx_*)\big) \\
    &\mu_N^{(i)}(\bfx_*) = K_i(\bfx_*, \bfx_{[N]})^\intercal (K_i(\bfx_{[N]}, \bfx_{[N]})+s^2I_N)^{-1} \bfy_{i,[N]} \\
    &\sigma_N^{(i)}(\bfx_*) =\kappa_i(\bfx_*, \bfx_*) - K_i(\bfx_*, \bfx_{[N]})^\intercal  (K_i(\bfx_{[N]}, \bfx_{[N]})+s^2I_N)^{-1} K_i(\bfx_*, \bfx_{[N]}),
\end{align}
where $K_i(\bfx_{[N]},\bfx_{[N]}) \in \mathbb{R}^{N \times N}$ with $[K_i(\bfx_{[N]},\bfx_{[N]})]_{j,k}=\kappa_i(\bfx_j,\bfx_k)$, and $K_i(\bfx_{*},\bfx_{[N]}) \in \mathbb{R}^{1 \times N}$ with $[K_i(\bfx_{*},\bfx_{[N]})]_{j}=\kappa_i(\bfx_*,\bfx_j)$.  
We estimate $d(\bfx)$ with the mean of the GP posteriors. That is, $\hat{d}(\bfx)=\mu_N(\bfx)$, where $\mu_N(\bfx):=[\mu_N^{(1)}(\bfx), \ldots, \mu_N^{(n)}(\bfx)]^\intercal$.

\subsection{Sample-dependent high confidence error bound}
In this section, we derive a high probability upper bound on the number of observations required to ensure that the estimate error $e(x)$ can be made smaller than some prescribed value $\psith$ within a neighborhood of radius $\rho$ around some given point $\tbf{o}$. 

We begin by stating an assumption on the sampling distribution of the agent, which formalizes the requirement that the agent can gather sufficient information within its neighborhood. This assumption is necessary for our main result of this section, Theorem~\ref{prop:sample_complexity}, as our goal is to get uniformly good estimates of $d$ at every point in the neighborhood.

\begin{assumption}
\label{assumption:sampling_distribution} 
We assume that when the agent is situated at some point $\tbf{o} \in \mc{X}$, it can draw samples in a neighborhood $\ball(\tbf{o}, \rho)$ around the point according to a sampling distribution $Q$ with support $\ball(\tbf{o}, \rho)$, which admits a density $q$ satisfying the property $\cl \leq q(x)$ for all $x \in \ball(\tbf{o}, \rho)$ for a positive constant $\cl>0$. Note that a special case of $Q$ is the uniform distribution which admits a constant density $q(x) = \frac{1}{\text{Vol}(\ball(\tbf{o}, \rho))}$.%
\end{assumption}

We can now state the main result of this section which provides a bound on the number of observations needed to ensure a uniformly good estimate of the model error function $d$. 
\begin{theorem}
\label{prop:sample_complexity}
Suppose the following conditions are satisfied: \begin{inparaenum}[1)] \item The model error $d$ in~\eqref{dynm1-7} is a sample from a zero-mean GP with the covariance function $\mc{K}$ satisfying Assumption~\ref{assump:covariance}, and \item The agent can make observations in its neighborhood according to a sampling distribution $Q$ satisfying Assumption~\ref{assumption:sampling_distribution}.\end{inparaenum}~Then the number of observations $N(\rho, \delta)$, drawn according to the sampling distribution $Q$, that are required by the agent in a ball of radius $\rho$ around some point $\textbf{o}$ to ensure that $\|e(x)\|_2 \leq \psi$ for all $x \in \ball(\textbf{o}, \rho)$ with probability at least $1-\delta$ is $\widetilde{\mathcal{O}}\left(  \max \left \{ 2 \sqrt{n}\psi^{-1},\;  \frac{\psi^{-2n/\omega}}{\cl^2}, \;  \frac{s^2 \psi^{-(2\omega+n)/\omega}}{\cl} \right\} \right)$ where the $\widetilde{\mc{O}}(\cdot)$ notation suppresses the poly-logarithmic factors of $\log(1/\delta)$ and $\log(1/\psi)$. 

\end{theorem}

\begin{proof}(sketch)
Full proof can be found in Appendix~\ifx\extended\undefined A of the extended paper~\citep{sun2020uncertainty}\else \ref{appendix:proof_sample_complexity}\fi.
\begin{itemize}
    \item First, following the proof of \citep[Lemma~5.6]{srinivas2009gaussian}, we first introduce a high probability event $\Omega_1$ such that for all points $x$ in a fine discretization~(denoted by $H$) of $\ball(\tbf{o}, \rho)$, we have $|\mu_t^{(j)}(x) - d^{(j)}(x)|\leq \beta_N \sigma^{(j)}_{t-1}(x)$ for $1 \leq j \leq n$. By making the discretization $H$ fine enough, we can ensure sufficiently accurate estimate of $d$ at every point of $\mc{X}$ by appealing to property~(iii) in Assumption~\ref{assump:covariance}. 
    \item Next, we note that by using \citep[Prop.~3]{shekhar2018gaussian}, to ensure uniformly tight estimate, we need to ensure that every point $x \in \ball(\tbf{o}, \rho)$  has sufficiently many samples in a ball of radius $r_0$ around it, for an appropriate choice of $r_0$. To achieve this, we consider a fixed $r_0/2$-covering of $\ball(\tbf{o}, \rho)$, denoted by $E$, and find $N$ large enough which ensures that a $r_0/2$ neighborhood of every point in $E$ has sufficiently many samples drawn according to $Q$. 
\end{itemize}
\end{proof}
\vspace{-0.5cm}
\begin{remark}
Note that our proof of Theorem~\ref{prop:sample_complexity} proceeds by first obtaining a uniform deviation bound by controlling the deviation on the elements of a sufficiently fine discretization $H$ of the ball $\ball(\textbf{o}, \rho)$. Alternatively, we could also have employed the uniform error bounds derived in \citep[Theorem~3.1]{lederer2019uniform} for this task. However, our approach leads to a slightly easier path to obtain the sample complexity, i.e., finding the value of $N$ which ensures that the error is smaller than some given quantity $\psi$. Performing this ``inversion" with the more general bounds derived by \cite{lederer2019posterior,gahlawat2020l1} may be more involved. 
\end{remark}

%% file: tracking_error.tex
\section{Learning-based tracking controller and tracking error}
\label{sec:C3M}
In the last section, we constructed  a high confidence bound on the estimation error of the disturbance. In this section, we show how to learn a tracking controller with high confidence bound on the tracking error.

Contraction theory~\citep{lohmiller1998contraction} analyzes the incremental stability of a system by considering the evolution of the distance between any pairs of arbitrarily close neighboring trajectories. The existence of a Control Contraction Metric (CCM)~\citep{manchester2017control} ensures the existence of a tracking controller that can drive the system to any nominal trajectories.

We apply the method proposed by~\cite{sun2020learning} to jointly learn a tracking controller and a contraction metric function for the dynamics with the estimate of the disturbance, i.e. $\dot{\bfx}=\hat{f}\left(\bfx(t)\right) + B\left(\bfx(t)\right)\bfu(t)$, where $\hat{f}(\bfx) = f(\bfx) + \hat{d}(\bfx)$. As shown by~\cite{sun2020learning}, the learned metric $M(\cdot)$ is just a mapping from the state $\bfx$ to an $n \times n$ positive definite matrix. The learned tracking controller is a feedback controller of the form $\bfu(\bfx,\bfx^*,\bfu^*)$, where $\bfx$ is the current state and $\bfx^*, \bfu^*$ are the nominal state and control input. We want to find a metric function $M(\cdot)$ and a feedback controller $\bfu(\cdot)$ satisfying that for all $\bfx \in \mathcal{X}$, $\bfx^* \in \mathcal{X}$, $\bfu^* \in \mathcal{U}$, and some $\lambda > 0$,
\begin{equation}
\label{eq:CCM_condition}
    \dot{M} + \sym{M (A+BK)} + 2 \lambda M \prec 0,
\end{equation}
where $A := \frac{\partial \hat{f}}{\partial \bfx} + \sum_{j=1}^{m} \bfu^{(j)} \frac{\partial b_j}{\partial \bfx}$,  $b_j$ is the $j$-th column of $B$,  $\bfu^{(j)}$ is the $j$-th element of $\bfu$, $K = \frac{\partial \bfu}{\partial \bfx}$, and $\dot{M}$ is the derivative of $M(\bfx(t))$ w.r.t. time. We refer the readers to~\citep{sun2020learning} for more details. Note that the above formulation uses the estimated dynamics by plugging $\hat{d}(\bfx)$ in~\eqref{dynm1-7}. The following theorem shows that when applied to the real dynamics, the tracking error of the learned controller is still bounded.
\begin{theorem}[Robustness to dynamics error,~\citealt{sun2020learning}]
\label{thm:robustness}
Given $M$ and $\bfu$ satisfying inequality~\eqref{eq:CCM_condition}, since $M(\bfx)$ is positive definite, there exist $\overline{m} \geq \underline{m} > 0$ such that $\underline{m}\mathbf{I} \preceq M(\bfx) \preceq \overline{m}\mathbf{I}$ for all $\bfx$. Assume that error of the dynamics is bounded as $\|e(\bfx)\| \leq \psi$ for all $\bfx$ and some $\psi >0$. Now considering the trajectory $\bfx(t)$ of the closed-loop system, the distance between $\bfx(t)$ and any given reference $\bfx^*(t)$ is bounded as $\|\bfx(t) - \bfx^*(t)\|_2 \leq \frac{R_0}{\sqrt{\underline{m}}} e^{-\lambda t} + \sqrt{\frac{\overline{m}}{\underline{m}}} \cdot \frac{\psi}{\lambda} (1 - e^{- \lambda t})$,
where $R_0$ is the Riemannian distance between $\bfx(0)$ and $\bfx^*(0)$ under metric $M$.
\end{theorem}

Moreover, if $\bfx(0) = \bfx^*(0)$, then the Riemannian distance $R_0 = 0$. This is usually the case since the reference trajectory planned by the open-loop motion planner exactly starts from the current state of the agent. Thus, the tracking error of the learned controller is upper bounded by $\te = \sqrt{\frac{\overline{m}}{\underline{m}}}  \frac{\psi}{\lambda}$. If we can ensure that the planned nominal trajectory is at least $\te$ away from the obstacles, then the realized trajectory is guaranteed to be safe. As will be shown in Sec.~\ref{sec:algo}, this is equivalent to bloating the obstacles by $\te$ before planning.
The following corollary immediately follows from Theorem~\ref{prop:sample_complexity} and Theorem~\ref{thm:robustness}.
\begin{corollary}
 Suppose that a ball $\ball(\bfo, \rho)$ contains $N$ samples such that $N$, $\rho$, $\delta$, $\psi$ satisfy the condition of Theorem~\ref{prop:sample_complexity} for some $\delta$ and $\psi$. If there exists a controller and metric satisfying the CCM condition~\eqref{eq:CCM_condition} and the motion of the closed-loop system is restricted in $\ball(\bfo, \rho)$, then the tracking error is less than or equal to $\te = \sqrt{\frac{\overline{m}}{\underline{m}}} \cdot \frac{\psi}{\lambda}$ with probability at least $1-\delta$.
\end{corollary}

\paragraph{Retraining of the controller.} As mentioned before, the agent gradually collects more and more observations and keeps improving the estimate $\hat{d}$. In this case, we might have to learn a new controller $\bfu$ and a new contraction metric $M$ such that condition~\eqref{eq:CCM_condition} still holds. Retraining of this controller is expensive, and thus we use the following method to reduce the number of retrainings. The basic idea is to impose some robust margin on condition~\eqref{eq:CCM_condition} during training, such that the learned metric and controller are robust to the change of $\hat{d}$ to some extent. Specifically, instead of condition~\eqref{eq:CCM_condition}, we use the following condition for learning,
\newcommand{\margin}{\mathcal{M}}
\begin{equation}
\label{eq:robust_CCM_condition}
    \dot{M} + \sym{M (A+BK)} + 2 \lambda M \prec -\margin\mathbf{I},
\end{equation}
where $\mathcal{M} > 0$ is the margin for robustness. Intuitively, if we impose the above condition, small changes in $\hat{d}$ will not lead to a violation of condition~\eqref{eq:CCM_condition}. Retraining is only needed when the change in $\hat{d}$ crosses a certain threshold. 
Formally, we have the following theorem.

\newcommand{\dres}{\mathcal{R}}
\begin{theorem}
\label{thm:retraining}
Consider two estimates $\hat{d}_1$ and $\hat{d}_2$ and their difference $\dres = \hat{d}_1 - \hat{d}_2$. If the metric $M$ and controller $\bfu$ satisfy the robust condition~\eqref{eq:robust_CCM_condition} for the estimate $\hat{d}_1$ and the difference $\dres$ satisfies the following condition for all $\bfx$,
\begin{equation}
\label{eq:retrain_condition}
\|\partial_{\dres}M + \sym{M\dres}\|_2 \leq \margin,
\end{equation}
then the original condition~\eqref{eq:CCM_condition} is also satisfied for the estimate $\hat{d}_2$.
\end{theorem}

The proof can be found in Appendix~\ifx\extended\undefined B of \citep{sun2020uncertainty}\else \ref{appendix:proof_retraining}\fi. In practice, we evaluate condition~\eqref{eq:retrain_condition} only in the region of our interest instead of the whole state space. Moreover, evaluating whether condition~\eqref{eq:retrain_condition} holds on an uncountable set is hard. Instead, we randomly sample a number of points from the set and say condition~\eqref{eq:retrain_condition} holds for the whole set only if it holds for all sampled points with a robust margin determined by the Lipschitz constant of the LHS of condition~\eqref{eq:retrain_condition}~\citep[cf.][Sec. 3.2]{sun2020learning}.

%% file: algo.tex
\section{Algorithm}
\label{sec:algo}
\begin{figure}[tbp]
\vspace{-1cm}
\noindent\begin{minipage}{.5\textwidth}
\begin{algorithm2e}[H]
\KwIn{Initial state $\bfx$; Obstacles $\mathcal{O} \subset \mathcal{X}$;}
\KwIn{Error tolerance $\psith$; Confidence level $\delta$;}
\KwOut{Final estimate $\hat{d}$;}
\SetKwFunction{FPlan}{Plan}
\SetKwProg{Fn}{Function}{:}{}
\Fn{\FPlan{$\bfx$, $\bfg$, $\te$}}{
    \KwData{current state $\bfx$; goal $\bfg$; bloating factor $\te$;}
    Bloating obstacles: $\tilde{\mathcal{O}} = \mathcal{O} \bigoplus \ball(0,\te)$\;
    Plan from $\bfx$ to $\bfg$ while avoiding $\tilde{\mathcal{O}}$\;
}

\While{not satisfied}{
    Find next goal $\bfg$ to visit using Eq.~\eqref{eq:next_point}\;
    $\rho = \rho_0$; $\mathtt{path} = \mathtt{null}$\;
    \While{$\mathtt{path}$ is $\mathtt{null}$}{
        Compute $\te$ in $\ball(\bfx, \rho)$\;
        $\mathtt{path}$ = \FPlan{$\bfx$, $\bfg$, $\te$}\;
        Decrease $\rho$\;
    }
    Move along $\mathtt{path}$ until\\ ~~~reaching the boundary of $\ball(\bfx, \rho)$\;
    Enlarge the observation set and update $\hat{d}$\;
    Retrain the controller if needed\;
}
\caption{Safe exploration.}
\label{alg:alg}
\end{algorithm2e}
\end{minipage}
\begin{minipage}{.5\textwidth}
\centering
\includegraphics[width=\textwidth]{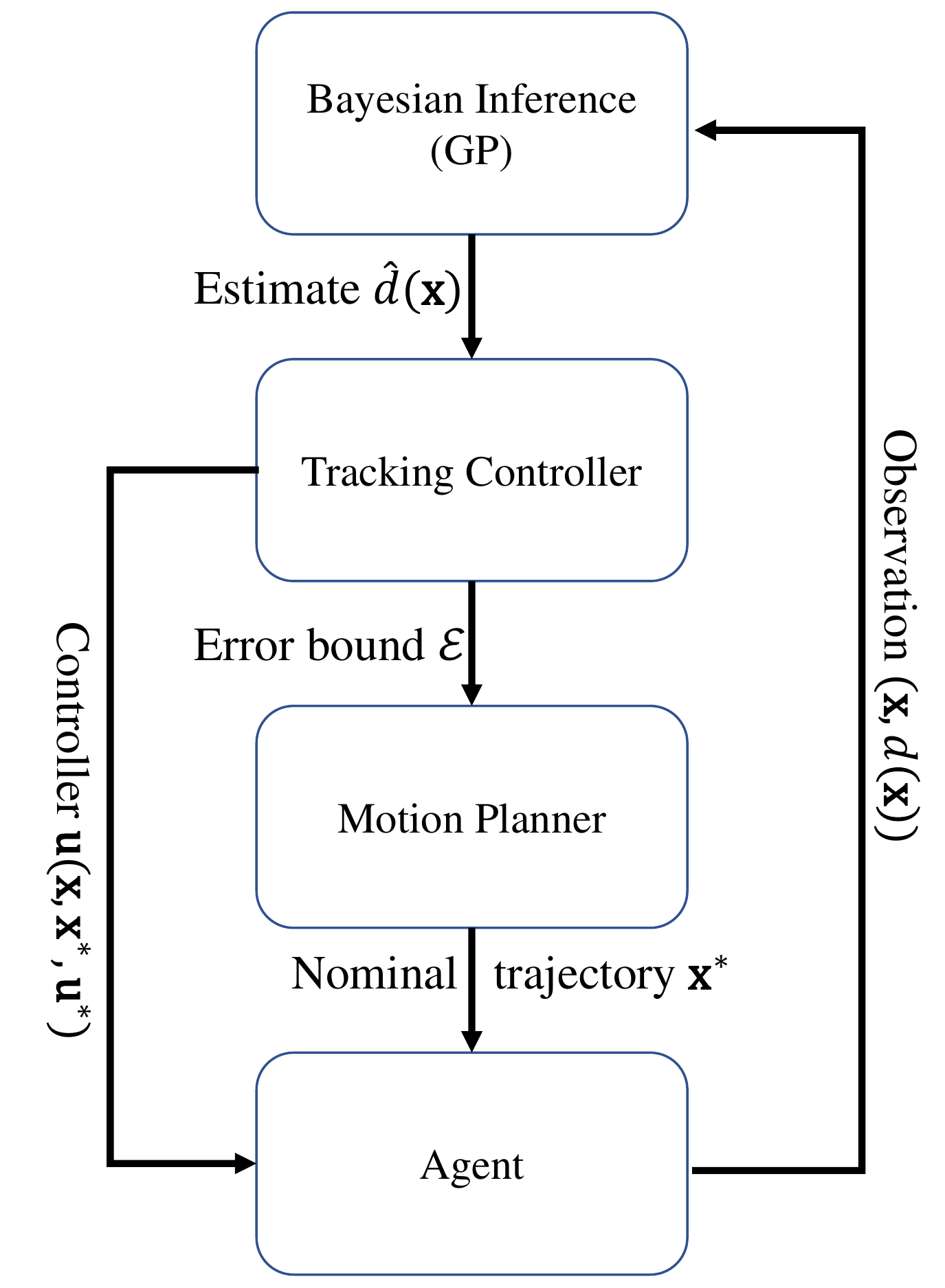}
\captionof{figure}{Diagram of the proposed method.} \label{fig:arch}
\end{minipage}
\end{figure}
The overall framework is shown in Algorithm~\ref{alg:alg}. Several components are explained in order.
\paragraph{Compute the estimation error.} 
In Algorithm~\ref{alg:alg}, we need to determine the estimation error $\psi$ in a ball $\ball(\bfo, \rho)$ given the current observations and the confidence level $\delta$.
 Theorem~\ref{prop:sample_complexity} provides a high probability bound on the number of samples needed to ensure uniformly good estimate within a ball $\ball(\textbf{o}, \rho)$.
Based on this theorem, we now present a practical heuristic to  compute an upper bound on estimation error, which in turn provides a stopping rule for the sampling. 
We proceed as follows. We use black-box optimization to find the maximizer of $\sigma^{(j)}_N$ over the domain $\ball(\textbf{o}, \rho)$ for all $1\leq j \leq n$ which we denote by $\tilde{\sigma}^{(j)}_N$. As we mentioned in the proof sketch of Theorem~\ref{prop:sample_complexity} the absolute value of the estimation error for the $j$-th coordinate is bounded by $\beta_N \sigma^{(j)}_{t-1}(x)$.
Hence, we stop the sampling if $\beta_N \sqrt{\sum_{j=1}^{n} [{\tilde{\sigma}^{(j)}_N}]^2}$, which represents the upper bound on the 2-norm of the total estimation error, is smaller that $\psith$. Here, $\beta_N$ is a quantity defined in Appendix~\ifx\extended\undefined A of \citep{sun2020uncertainty}\else \ref{appendix:proof_sample_complexity}\fi. %
\paragraph{Find the next point to visit.} At each time step, the agent has to determine the next point to visit and collect observations around its trajectory. To make the exploration efficient, the next point to visit must be informative. Therefore, we choose the one with highest estimate variance. Formally,
\begin{equation}
\label{eq:next_point}
    \bfg = \arg\max_{\bfx \in \mathcal{X}} \sum_{i=1}^{n}\left( \kappa_i(\bfx, \bfx) - K_i(\bfx, \bfx_{[N]})^\intercal  (K_i(\bfx_{[N]}, \bfx_{[N]})+s^2I_N)^{-1} K_i(\bfx, \bfx_{[N]})\right).
\end{equation}
\paragraph{Planning a feasible path.} After determining the next point to visit, the agent has to plan a nominal trajectory feasible for the nominal dynamics (i.e. without considering $e(\bfx)$) such that controlled by the learned controller in Sec.~\ref{sec:C3M}, the agent can safely track this nominal trajectory and reach the goal. As mentioned in Sec.~\ref{sec:C3M}, the distance between the actual trajectory and the nominal one is bounded by $\te$. Therefore, we first bloat the obstacles with the error bound $\te$: Let the obstacles be $\mathcal{O} \in \mathcal{X}$; the bloated obstacles are just $\mathcal{O} \bigoplus \ball(0,\te)$, where $\bigoplus$ denotes the Minkowski summation and $\ball(0,\te)$ denotes the ball centered at the origin with radius $\te$. Then, a nominal trajectory to the goal is planned while avoiding the bloated obstacles. Any motion planner could suffice, e.g.~\citep{vitus2008tunnel,fan2020fast}, and we adopt RRT*~\citep{karaman2011sampling}. RRT* is used to generate a piece-wise linear path. However, this path is usually not feasible for the agent. Thus, an additional step is required to generate a feasible trajectory and the corresponding reference control input. To this end, a simple linear feedback controller is used to track the planned piece-wise linear path (again, without considering $e(\bfx)$). The trajectory generated by the agent controlled by the simple controller will be used as the reference, i.e. $\bfx^*(t)$ and $\bfu^*(t)$ for the tracking controller. Due to the tracking error introduced by the simple controller, $\bfx^*$ may be unsafe. If that happens, we will bloat the obstacles a bit more and repeat planning until we find a safe $\bfx^*$. However, in the experiments we found that this was very rarely needed.
\paragraph{Putting it all together.} At each time step, the agent first determines the next point to visit. Then, it initializes the radius $\rho = \rho_0$ and computes the upper bound on the estimation error  and the corresponding tracking error $\te$ in the ball $\ball(\bfx, \rho)$. Using $\te$, the agent searches for a safe path to the goal. If it failed to find such a path, then $\rho$ is decreased a bit and the above process is repeated until a safe path is found. Then, controlled by the learned controller, the agent moves along the path and collects new observations on the disturbance until it reaches the boundary of the ball $\ball(\bfx, \rho)$. Then, GP is invoked to update the estimate $\hat{d}$ using the new observations. After that, we might retrain the controller if needed as shown in Sec.~\ref{sec:C3M}. The exploration will terminate once we have collect enough samples such that $\|e(\bfx)\|_2 \leq \psith$ for all $\bfx \in \mathcal{X} \setminus \mathcal{O}$ with probability at least $1-\delta$.

%% file: experiments.tex
\section{Experimental evaluations}
In order to evaluate the proposed safe-exploration framework, we designed a scenario as shown in Fig.~\ref{fig:scene}. Several components of the scenario are explained in order.
\paragraph{Dynamics.} We adopted the Dubins car model for the agent. The state of the system is $\bfx := [p_x, p_y, \theta, v, \omega]^\intercal$, where $(p_x, p_y)$ the position of the car, $\theta$ the heading angle, $v$ the velocity, and $\omega$ is the angular velocity. The control input is $\bfu := [f, \tau]^\intercal$, where $f$ is the force and $\tau$ is the torque. The dynamics of the car is
\[
\dot{\bfx} = \begin{bmatrix}v\cos(\theta)\\v\sin(\theta)\\\omega\\-0.4 v\\-0.4 \omega\end{bmatrix} + \begin{bmatrix}0 & 0\\ 0 & 0\\ 0 & 0\\ 1 & 0\\ 0 & 1\end{bmatrix} \bfu + d(\bfx).
\]
\paragraph{Workspace and obstacles.} In the experiments, we want the agent to explore a square region $[0, 10] \times [0,10]$ on the 2D plane. We randomly generate $10$ obstacles, which are shown in Fig.~\ref{fig:scene}.
\paragraph{Disturbance function.} The disturbance $d(\bfx)$ is a function of the first two elements of $\bfx$ and models the effect caused by the ground at position $(p_x, p_y)$. We use an image from~\citep{sturtevant2012benchmarks} as the terrain map. In order to define the $5$-dimensional disturbance based on the color values of the image, we use a $5 \times 3$ projection matrix $P$ to map the color space to disturbance space. The disturbance at $(p_x, p_y)$ is obtained by multiplying $P$ and the corresponding RGB color value.
\paragraph{Simulation.} At the beginning of the simulation, we assign a random initial position to the agent such that it is safe initially. Then, the trajectories are simulated with a constant time step $\Delta t = 0.01$~s. At each time step, we check the safety of the agent. If the distance to an obstacle is less than a threshold $\thr > 0$, then it is said to be unsafe. In the following experiments, we set $\thr$ to $0.1$ meters. The goal of the agent is to collect observations on the disturbance to construct an estimate $\hat{d}$ of the actual disturbance map $d$ and maintain safety in this process.
\paragraph{Metrics for evaluation.} We use the following metrics for comparison: 1. {\em Unsafe} is the percentage of iterations at which the agent is unsafe; 2. {\em Travel time} is the time needed for exploring the workspace; 3. {\em Tracking error} is the average tracking error, which is the distance between the actual trajectory and its nominal trajectory averaged over all time steps. Moreover, all the metrics reported in this section are averaged over $5$ runs.
\paragraph{Comparison with the baseline method.} The baseline method is a variant of Algorithm~\ref{alg:alg}. The baseline method does not make use of the current estimate $\hat{d}$ to retrain the controller and compute the high-probability tracking error $\te$. Instead, $\te$ is set to be a constant. In the experiments, we set $\te = 0$, $0.1$, or $0.3$. We also tried to use larger $\te$, e.g. $\te=0.6$, however, in that case, the bloated obstacles blocked the free space, which makes it impossible to finish the exploration. For all the methods we set $\rho_0 = 1$, $\psith = 0.1$, and $\delta = 0.05$. The results are shown in Table~\ref{tab:results}. Compared to the baseline methods, the proposed method results in higher safety and shorter travel time, which demonstrates the sufficiency of the proposed method.
Further illustration of the experimental results can be found in Appendix~\ifx\extended\undefined C of \citep{sun2020uncertainty}\else\ref{appendix:moreexp}\fi.

\begin{table}[tbp]
    \centering
    \caption{Comparison with the baseline method.}
    \begin{tabular}{|c|c|c|c|}
        \hline
        Method & Unsafe ($\%$) & Travel time (s) & Tracking error\\\hline
        Algorithm~\ref{alg:alg} & 0.3 & 236 & 0.051 \\\hline
        Baseline ($\te=0$) & 10.3 & 208 & 0.243 \\\hline
        Baseline ($\te = 0.1$) & 5.1 & 314 & 0.221\\\hline
        Baseline ($\te = 0.3$) & 4.0 & 515 & 0.230\\\hline
    \end{tabular}
    \label{tab:results}
\end{table}

%% file: appendix_proof_sample_complexity.tex
To present the details of the proof, we need to introduce some additional notation. Let $E$ denote the $r_0/2$ covering of $\ball(\tbf{o}, \rho)$ for some $0<r_0<1$, and $H$ denote the $r_1$ covering for some $r_1<r_0/2$. Both the terms $r_0$ and $r_1$ will be specified later. Throughout this proof, we will use $m_E$ and $m_H$ to denote the cardinality of $E$ and $H$ respectively, and furthermore, for any $x \in \ball(\tbf{o}, \rho)$ we will use $[x]_E$ and $[x]_H$ to denote the element in $E$ and $H$ (respectively) that is closest to $x$. In the case of more than one point being the closest we will choose according to some predetermined rule. 
Finally, we will enumerate the elements of $E$ as $\{z_1, z_2, \ldots, z_m\}$. 

Now, suppose that the agent draws $N$ i.i.d. points according to a sampling distribution $Q$ from the region $\ball(\textbf{o}, \rho)$, and denote the drawn points by $S_N = \{X_1, X_2, \ldots, X_N\}$.  Introduce the random variables $m_i = |S_N \cap \ball(z_i, r_0/2)|$, denoting the numbers of random samples falling in the $r_0/2$ neighborhood of $z_i$, for $1 \leq i \leq m$. 

For some given confidence level $\delta \in (0,1)$ we introduce the following three events which can be ensured to occur simultaneously with probability at least $1-\delta$. 
\begin{itemize}
    \item  Suppose the set $H$ is an $r_1 = 1/(NL\sqrt{n})$~(where $L = a_2 \sqrt{ \log (3a_1n/\delta)}$) covering of $\ball(\tbf{o}, \rho)$ (recall that the terms $a_1$ and $a_2$ from Assumption~\ref{assump:covariance}).   Introduce the event 
    \begin{align}
    \label{eq:event_Omega_1}
    \Omega_1 = \{ |d^{(j)}(z) - \mu^{(j)}(z)|\leq \beta_N \sigma_t^{(j)}(z),\; \forall z \in H, \forall 1 \leq t \leq N \},
     \end{align}
    where 
    \begin{align}
    \label{mmemeebeta}
    \beta_N = \sqrt{ 2\log (3Nm_H/\delta)} \text{ and } m_H = C_n\left(\frac{NL\sqrt{n}}{\rho}\right)^n
    \end{align}
      for some constant $C_n>0$ depending only on $n$. Then, we have $\mbb{P}(\Omega_1) \geq 1-\delta/3$
    
    \begin{proof}
    The proof of this statement proceeds along the lines of the proofs of \citep[Lemmas~5.5~\&~5.6]{srinivas2009gaussian}. In particular,  we note that for any $z \in H$, the posterior is a normal random variable with mean $\mu_t(z)$ and variance $\sigma_t^2(z)$, and thus by the Gaussian tail inequality and two union bounds (one over the elements of $H$ for a fixed $t$, and the second over $t=1,2,\ldots, N$) we get the required result.  
    \end{proof}
    
    \item Next, we introduce the event $\Omega_2 = \{ |\partial d(x)/\partial x| < L, \; \forall x \in \ball(\tbf{o}, \rho), \, \forall j=1,2,\ldots,n\}$ with $L= a_2 \sqrt{ \log \lp \frac{3a_1 n}{\delta} \rp }$. Then we have $\mbb{P}\lp \Omega_2\rp \geq 1- \delta/3$. 
    
    \begin{proof}
     This result follows directly from the assumption on the covariance function, stated in Assumption~\ref{assump:covariance}, that there exist constants $a_1$ and $a_2$ such that for any $L>0$, the event $\Omega_2$ is satisfied with probability at least $1- a_1 n e^{-L^2/a_2^2}$. The result then follows by plugging in the value of $L$ used in the definition of the event $\Omega_2$. 
    \end{proof}
    
    \item Finally, we introduce the event $\Omega_3 = \{ |m_i - Np_i| \leq \sqrt{2N \log (3m/\delta)}, \; \forall 1 \leq i \leq m \}$ where $p_i = \int_{\ball(z_i, r_0/2)} q(x)dx$ is the probability that a uniformly drawn sample from $\ball(\textbf{o}, \, \rho)$ falls in $\ball(x, r_0/2)$. Then we have $\mbb{P}\lp \Omega_3\rp \geq 1- \delta/3$. 
    
    \begin{proof}
    The result follows by an application of  Hoeffding's inequality and a union bound over elements of $E$ followed by another union bound over the $N$ time steps.
    \end{proof}
\end{itemize}
For the rest of the proof, we will work under the event $\Omega_1 \cap \Omega_2 \cap \Omega_3$,  which as shown above occurs with probability at least $1-\delta$. 

As a consequence of the simultaneous occurrence of $\Omega_1$ and $\Omega_2$, we note that for any $x \in \ball(\tbf{o}, \rho)$ we must have $d^{(j)}(x) \leq \mu_t^{(j)} \lp [x]_H\rp + \beta_N \sigma_t^{(j)}\lp [x]_H\rp + 1/N$.  Thus if $N \geq 2 \sqrt{n}/\psi$, then 
to obtain the required result, it suffices to show that $\beta_N \sigma_t^{(j)}(x) \leq \psi/(2\sqrt{n})$ for all $x \in H$. We proceed in the following  steps: 
\begin{itemize}
\item For any point $x \in H$, we note that there exists at least one $z_i \in E$ such that $\|x - z_i\|\leq r_0/2$. Consequently, the ball $\ball(z_i, r_0/2)$ is contained in the larger ball of radius $r_0$ centered around $x$, i.e., $\ball(x, r_0)$. Since, we assume that the event $\Omega_3$ holds, this implies that the number of random points from $S_N$ which fall in the ball $\ball(x, r_0)$ is at least $m_i \geq N \lp p_{r_0}- \sqrt{ \frac{2 \log(2m/\delta)}{N}}\rp$. 

Thus by an application of \citep[Proposition~3]{shekhar2018gaussian}, we note that after collecting $N$ observations, the approximation error at the point $x$ can be upper bounded as $|d^{(j)}(x) - \mu_t^{(j)}(x)|\leq \beta_N \sigma_t^{(j)}(x) \leq \beta_N \lp \frac{\sigma}{\sqrt{m_i}} + C_K r_0^\omega \rp$, where $C_K $ is introduced in Assumption~\ref{assump:covariance}.

Now, assuming that \textbf{(i)} $\beta_N \leq a$ for some $a >0$, and \textbf{(ii)} that $N$ is large enough to ensure that $\sigma/\sqrt{m_i} \leq C_K r_0^\omega$. Together these two assumptions imply that a suitable value of $r_0$ is $\lp \frac{\psi}{2a C_K } \rp^{1/\omega}$.

\item Now, we obtain the sufficient conditions on $N$ to ensure that the above two assumptions are satisfied. Recall, that we have already imposed the condition that $N$ is large enough to ensure that $1/N < \psi/(2\sqrt{n})$ or equivalently $N > 2 \sqrt{n}/\psi$. Additionally, we need $N$ to be large enough to ensure that $\sigma/\sqrt{m_i} \leq C_K r_0^\omega$, and we break it into two parts: 
\begin{itemize}
\item $N$ is large enough to ensure that $2 \log(3m/\delta)/N \leq (p_i/2)^2$, a sufficient condition for which is to ensure that $2 \log(3m/\delta)/N \leq (1/4)(\cl C_n r_0^n)^2$, where the term $\cl$ is introduced in Assumption~\ref{assumption:sampling_distribution}. Since $a^2 \geq 2 \log(2m/\delta)$ a sufficient condition for this is 
\begin{equation}
\label{eq:cond1}
N  \geq  a^{2 + 2n/\omega} \lp \frac{2 C_K}{\psi} \rp^{2n/\omega} \lp \frac{2^{2n-2}}{\cl^2 C_n^2}\rp
\end{equation}

\item $N$ is large enough to ensure that $\sigma/\sqrt{N p_i/2} \leq C_K r_0^\omega$ for all $i$, a sufficient condition for which is 
\begin{equation}
    \label{eq:cond2}
    N \geq \frac{ 2\sigma^2  }{C_K^2 C_n \cl} \lp \frac{2C_K a}{\psi}\rp^{(2\omega + n)/\omega}. 
\end{equation}
\end{itemize}

\item Now, it remains to show that there exists an $a>0$ such that if $N$ satisfies the above two conditions then $2 \log (2N^2/\delta) \leq a$. A sufficient condition for this is that 
\begin{align}
\label{eq:a_value}
    & a \geq 2\max \left \{ 
    \log \lp  \frac{ 8 \sigma^2 (2\rho)^{2n}}{\delta C_K^2} \lp \frac{C_K}{\psi}\rp^{(2\omega + n)/\omega} \rp, \; 
    \log \lp \frac{2 (2\rho)^{4n}}{\delta} \lp \frac{ C_K}{\psi} \rp^{4n/\omega} \rp, \; a^*
    \right\}, \text{with} \\
   & a^* =  \max \left \{ e^{-W\lp -1/( 8n/\omega + 8)\rp}, \; e^{-W \lp -\omega/(4\omega + 2n) \rp} \right \}, \quad \text{where } W \text{ is the Lambert W-function.}
\end{align}

\end{itemize}
To conclude, a sufficient condition for ensuring that the estimated value of $d$ is good enough with probability at least $1-\delta$ is that the agent draws at least $N$ uniform samples in the ball $\ball(\textbf{o}, \,\rho)$, where $N$ satisfies: 
\begin{align}
    \label{eq:N_value}
    N  = \widetilde{\mathcal{O}}\lp 
      \max \left \{ 2 \sqrt{n}\psi^{-1},\;  \frac{\psi^{-2n/\omega}}{\cl^2}, \;  \frac{s^2 \psi^{-(2\omega+n)/\omega}}{\cl} \right\}
    \rp, 
\end{align}
where the notation $\widetilde{\mc{O}}$ suppresses the polylogarithmic factors of $\log(1/\delta)$ and $\log(1/\psi)$ (arising from the conditions on $a$).

%% file: appendix_proof_retraining.tex
The following lemma is used for the proof of Theorem~\ref{thm:retraining}.
\begin{lemma}\label{lemma:eigdiff}
For any two symmetric matrices $A,B \in \reals^{n \times n}$, the difference of their largest eigenvalues satisfies:
\[
| \lambda_{\max}(A) - \lambda_{\max}(B)| \leq \|A - B \|_2.
\]
\end{lemma}

Lemma~\ref{lemma:eigdiff} is a well-known result that follows from the Courant-Fischer
minimax theorem. The detailed proof can be found at~\cite{fan2015bounded}.

\begin{proof} (of Theorem~\ref{thm:retraining}).
Plugging $\hat{d}_1$ and $\hat{d}_2$ into Equation~\eqref{eq:CCM_condition}, denote the LHS by $LHS(\hat{d}_1)$ and $LHS(\hat{d}_2)$ respectively. Then, we have
\[
LHS(\hat{d}_1) - LHS(\hat{d}_2) = \partial_{\dres}M + \sym{M\dres}.
\]
Then, following from Lemma~\ref{lemma:eigdiff} and the assumption that $\hat{d}_1$ satisfies the robust condition, we have
\begin{align*}
& \lambda_{\max}(LHS(\hat{d}_2))\\
\leq & \lambda_{\max}(LHS(\hat{d}_1)) + \|LHS(\hat{d}_1) - LHS(\hat{d}_2)\|_2\\
\leq & -\margin + \|\partial_{\dres}M + \sym{M\dres}\|_2\\
\leq & 0.
\end{align*}
Thus, $LHS(\hat{d}_2) \prec 0$, which means $\hat{d}_2$ satisfies the original condition~\eqref{eq:CCM_condition}.
\end{proof}

%% file: appendix_moreexp.tex
\begin{figure}
    \centering
    \includegraphics[width=0.45\textwidth]{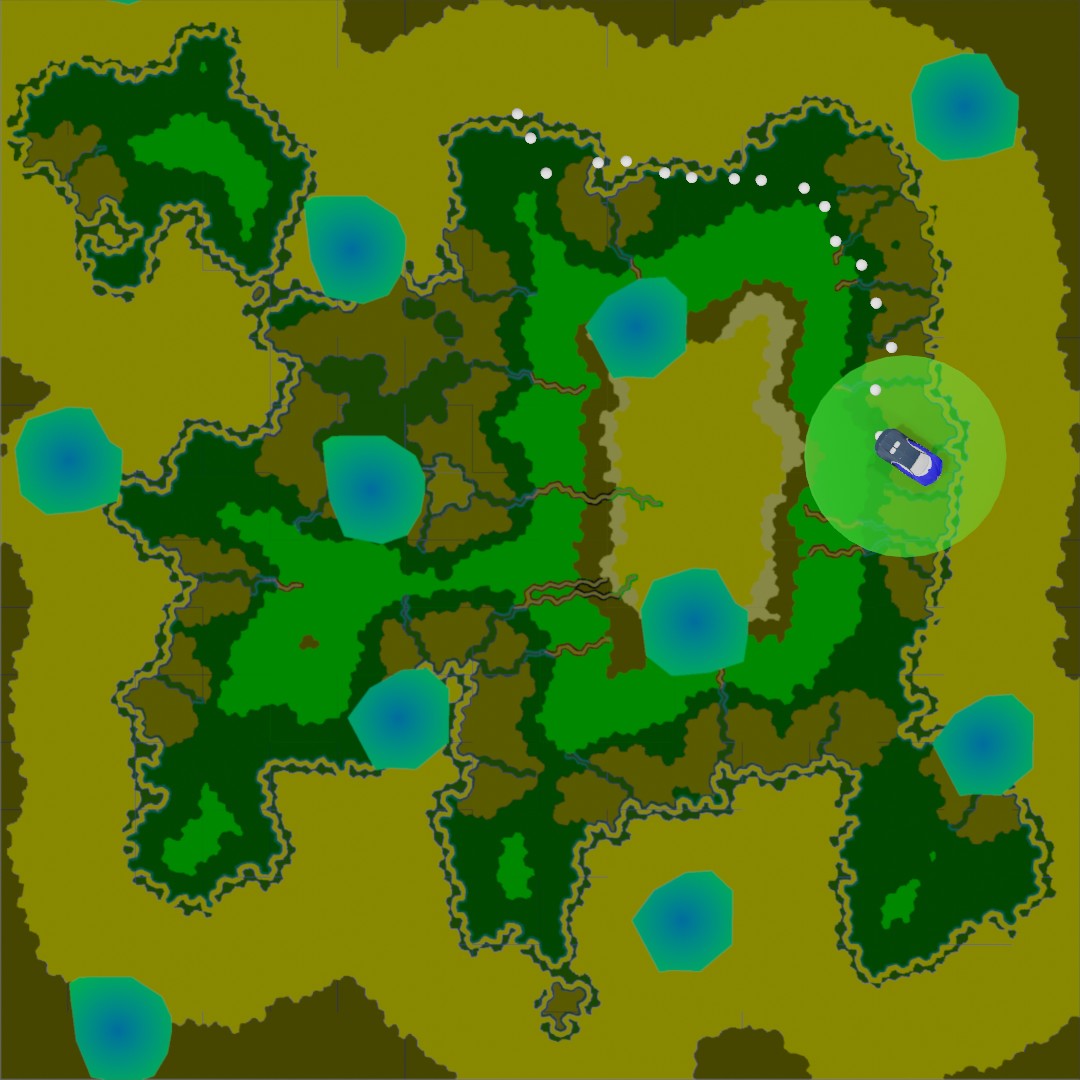}
    \includegraphics[width=0.45\textwidth]{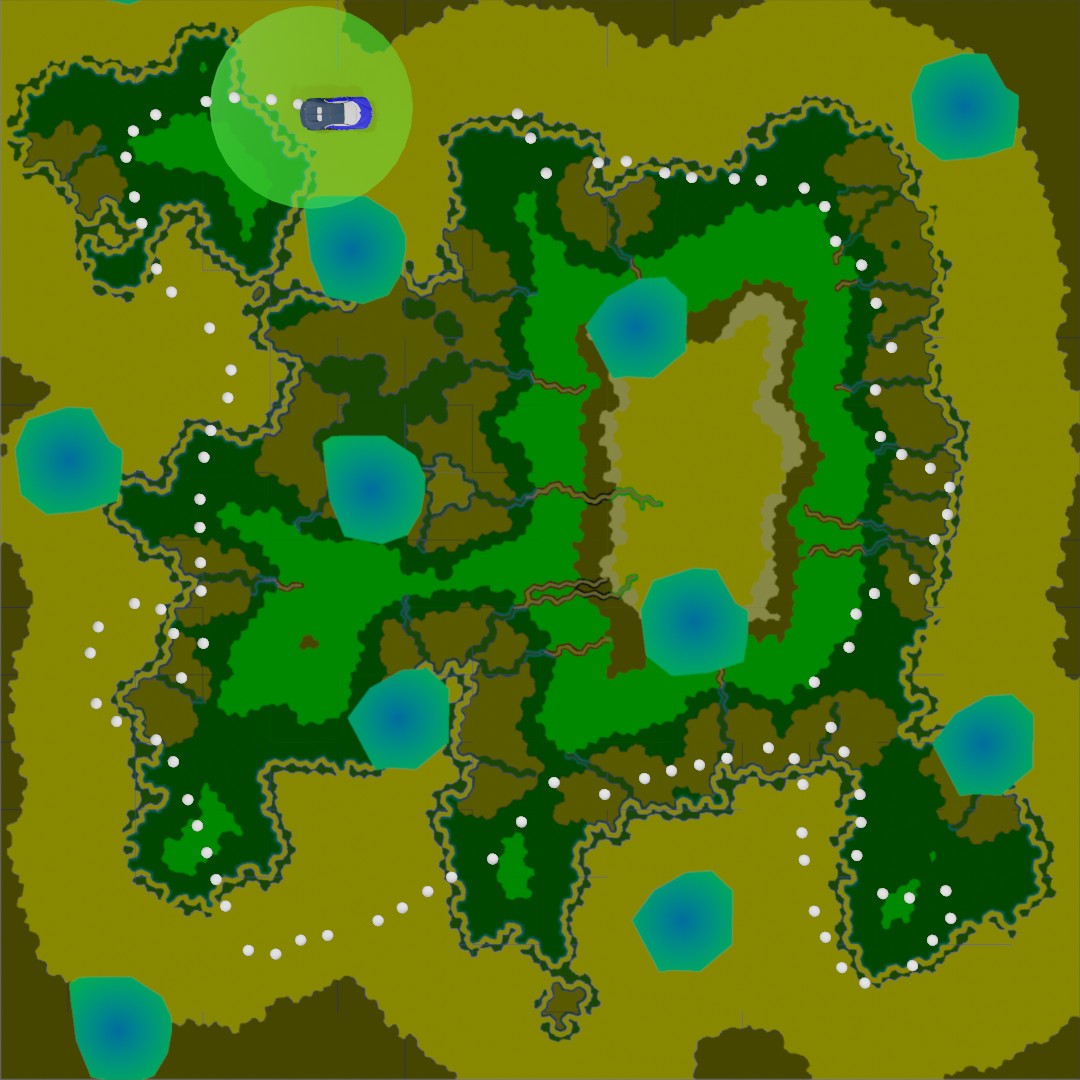}
    \includegraphics[width=0.45\textwidth]{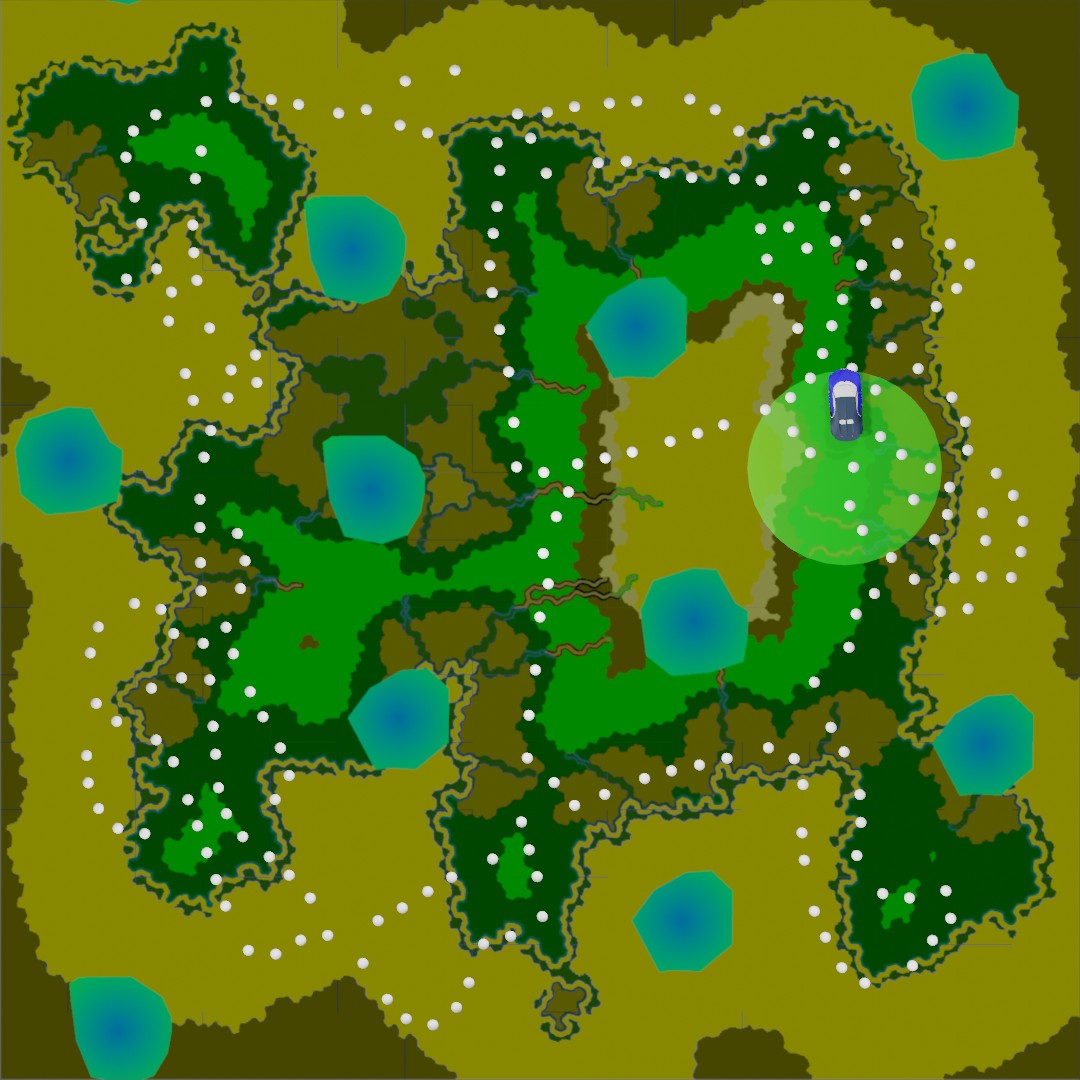}
    \includegraphics[width=0.45\textwidth]{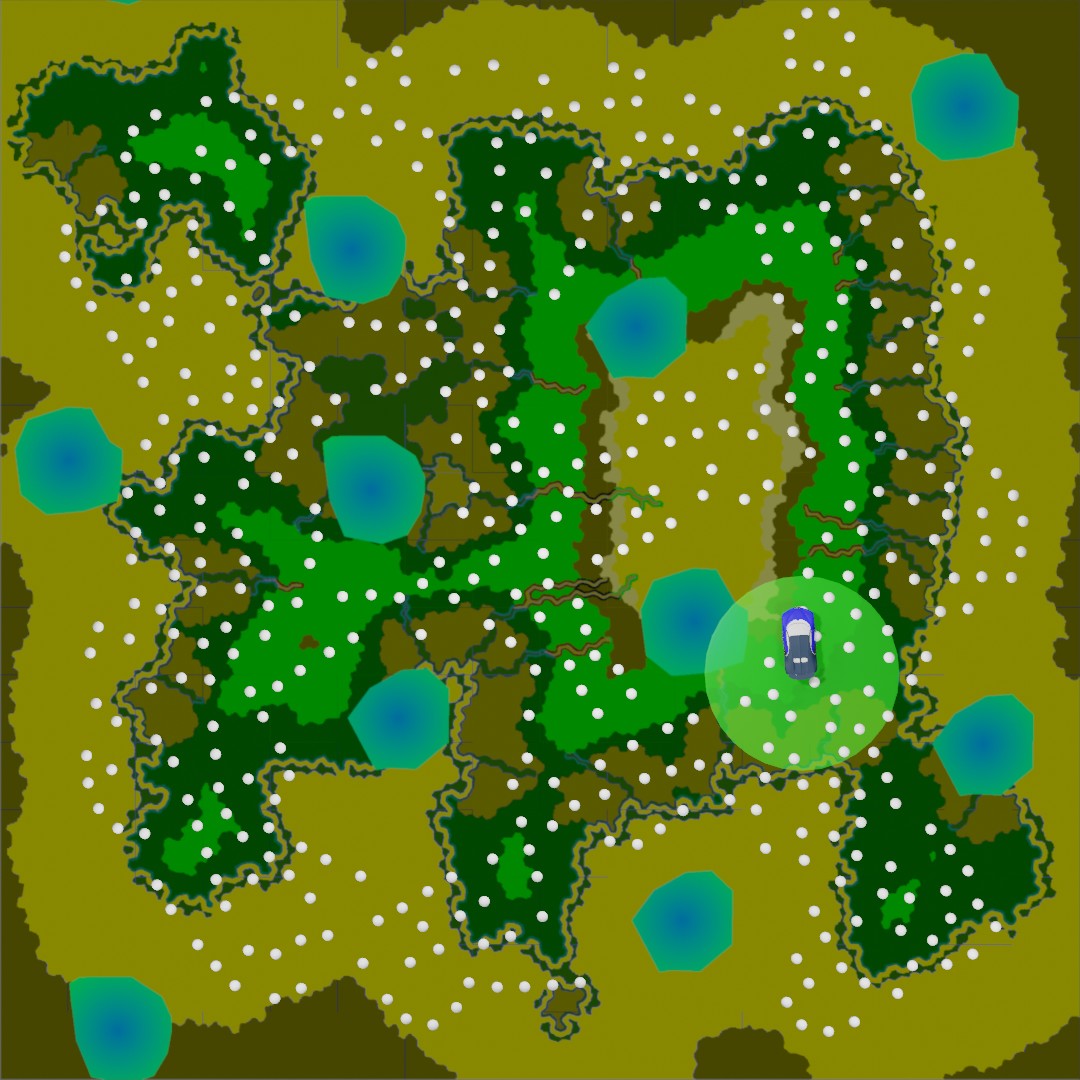}
    \caption{Exploration progress of the proposed method. White dots indicate the collected observations on the disturbance. Green transparent circle around the car is the ball $\ball(\bfx,\rho)$ in Algorithm~\ref{alg:alg}.}
    \label{fig:progress}
\end{figure}

The progress of exploration is visualized in Fig.~\ref{fig:progress}. A video is available at \url{https://youtu.be/cG4o29ntBbE}.

